\documentclass[manuscript]{acmart}

\renewcommand\footnotetextcopyrightpermission[1]{} 
\usepackage{booktabs} 
\usepackage[tight,normalsize,sf,SF]{subfigure} 
\usepackage{ epsf, float, graphicx, graphics}
\usepackage{setspace}
\usepackage{graphics,array}
\usepackage{bm}
\usepackage{blindtext}

\usepackage{xpatch}

\makeatletter
\xpatchcmd{\ps@firstpagestyle}{Manuscript submitted to ACM}{}{\typeout{First patch succeeded}}{\typeout{first patch failed}}
\xpatchcmd{\ps@standardpagestyle}{Manuscript submitted to ACM}{}{\typeout{Second patch succeeded}}{\typeout{Second patch failed}}    \@ACM@manuscriptfalse
\makeatother

\renewcommand\footnotetextcopyrightpermission[1]{} 
\setcopyright{none}
\DeclareMathAlphabet\mathbfcal{OMS}{cmsy}{b}{n}
\usepackage{etoolbox}
\setcopyright{none}

\begin{document}
\title{Mitigating Channel-wise Noise for Single Image Super Resolution}

\author{Srimanta Mandal}
\authornote{The work has been done during his post-doctoral studies at IPCV Lab, IIT Madras. The refined and complete version of this work appeared in the proceedings of the Indian Conference on Computer Vision, Graphics and Image Processing 2018 (ICVGIP'18).}
\orcid{0000-0003-3871-6621}
\affiliation{%
  \institution{DA-IICT}
  \city{Gandhinagar}
  \state{Gujarat}
  \postcode{382007}
}
\email{in.srimanta.mandal@ieee.org}

\author{Kuldeep Purohit}
\orcid{0000-0002-6709-1627}
\affiliation{%
  \institution{IIT Madras}
  \city{Chennai}
  \state{Tamil Nadu}
  \postcode{600036}
}
\email{kuldeeppurohit3@gmail.com}

\author{A. N. Rajagopalan}
\orcid{0000-0002-0006-6961}
\affiliation{%
  \institution{IIT Madras}
  \city{Chennai}
  \state{Tamil Nadu}
  \postcode{600036}
}
\email{raju@ee.iitm.ac.in}

\renewcommand{\shortauthors}{S. Mandal et al.}

\begin{abstract}
{\bf ABSTRACT}\\
In practice, images can contain different amounts of noise for different color channels, which is not acknowledged by existing super-resolution approaches. In this paper, we propose to super-resolve noisy color images by considering the color channels jointly. Noise statistics are blindly estimated from the input low-resolution image and are used to assign different weights to
different color channels in the data cost. Implicit low-rank structure of visual data is enforced via nuclear norm minimization in association with adaptive weights, which
is added as a regularization term to the cost.
Additionally, multi-scale details of the image are added to the model through another
regularization term that involves projection onto PCA basis, which
is constructed using similar patches extracted across different scales of the input image. The results demonstrate the super-resolving capability of the approach in real scenarios.

\end{abstract}


\maketitle

\section{Introduction}
Imaging systems are susceptible to noise in different conditions. 
Low light imaging requires higher analog gain (ISO value) of a camera, responsible 
for noise inclusion in the imaged scene~\cite{Bench_cvpr17}.
Now-a-days, majority of the images are captured by smartphone cameras
as compared to the point-and-shoot and DSLR cameras. 
However, the ease of photography
with smartphone often comes with the cost of higher levels of noise owing to smaller size of sensors~\cite{Brown_cvpr18}. Further, the captured images
can be lower in resolution, which may not fulfill the requirement of different HD applications. Thus, it becomes necessary to recover a noise less high resolution 
(HR) image from the captured noisy low resolution (LR) image. 

The LR image formation process can be represented mathematically as
\begin{eqnarray}
\label{eq:SR_prob}
\mathbf{y = DHx+n},
\end{eqnarray}
where $\mathbf{y}\in \Re^{N}$ is the LR observation, which is generated from the blurred (by $\mathbf{H} \in \Re^{M \times M}$ matrix) and decimated (by $\mathbf{D} \in \Re^{N \times M}$ matrix) version of the HR scene $\mathbf{x}\in \Re^{M}$ with
an additive noise $\mathbf{n}\in \Re^{N}$ ($M>N$). This mathematical model is devised from the imaging
pipeline, where decimation happens due to limited size of the sensors~\cite{overview}. 
The objective of super resolution (SR) is to achieve an estimation of $\mathbf{x}$
from $\mathbf{y}$, and is an ill-posed one as the number of unknowns ($M$) exceeds the number of equations ($N$). Further, the presence of $\mathbf{n}$ increases the perplexity of the problem.

The ill-posed objective can be partially accomplished using existing super resolution 
(SR) techniques~\cite{POCS,xample,ASDS,yang,Mandal_SSD}. As the high frequency (HF) information often gets attenuated or degraded in the imaging process, the SR approaches involve incorporating high frequency information. This is generally imported either 
from sub-pixel shifted multiple LR target images or from example images with HF content. Further, the ill-posed nature is often subjugated by prior information such
as total-variation, non-local similarity, sparsity, etc.~\cite{TVSR,ASDS,Mandal_SRSI,Mandal_SSD}. Absence of example image database or 
multiple LR images of the target scene with sub-pixel shift criteria can make these
kinds of SR approaches paralyzed. Such scenario can be addressed by utilizing the 
intra/inter-scale patch similarity~\cite{Glasner,self_similar,epsr3,Icvgip14,SISR_noise,Mandal_SRSI}. In the inverse
process, presence of noise often plays as a malefactor. Only a few approaches of SR are reported in the literature that works in noisy situation~\cite{SISR_noise,Mandal_SRSI,NNM_SR}. Furthermore, 
unknown statistics of noise can make the restoration more difficult~\cite{Mandal_SRSI,NNM_SR}. 

The existing approaches consider only the luminance component for SR by neglecting
the color information~\cite{elad,ASDS,Aplus,Mandal_SRSI,NNM_SR}. However, in real 
scenario, noise can be present in different amounts in different color channels~\cite{Noise_est_freeman,demos_noisy,MC-WNNM,CCINM,CIP}. This is because different channels have different ISO sensitivities. Further, the relative sensitivities vary with different WB settings.
Thus, super-resolving the luminance component may not be able to handle the
channel varying noise. One strategy could be to apply SR algorithms separately on each of the color channels. However, distinct processing of each color channel disregards the correlation among channels. 

In this paper, we propose to super-resolve a real noisy color image by considering the
color channels jointly to explore the correlation among the color channels.
Further, different weights are assigned to different color channels in the data 
cost in order to address the channel varying noise. The weights are estimated using the 
noise statistics from each channel. 
The low-rank property of clean data is approximated by incorporating
weighted nuclear norm.
Though, the nuclear norm
minimization strategy has been employed in SR by the work~\cite{NNM_SR}, it minimizes the nuclear norm uniformly without considering the significance of different singular values. Further, the approach~\cite{NNM_SR} super-resolves the luminance component only.
Whereas, we consider the significance of different singular values by adaptively weighting them, as weighted nuclear norm minimization can restore an image better than  uniform minimization~\cite{WNNM}. Moreover, the spectral correlation is utilized by
considering all the color channels jointly with adaptive weights. 
Multi-scale image details are embraced in the formulation by augmenting
another regularization term that involves projection onto PCA basis, learned from inter-scale similar RGB patches.

Here our main assumption is that the noise of real color image in standard RGB (sRGB) space can be approximated by the multivariate Gaussian model, as demonstrated in~\cite{CCINM}. Other noise such
as Poisson may not be well suppressed by our approach, however
presence of other additive noise can be adequately addressed.

Rest of the paper is sequenced as follows: Section~\ref{sec:related} discusses some of 
the related works, and highlights the contributions. The noise statistics are analyzed for real color images in Section~\ref{sec:proposed}, where the proposed approach of weighted data 
cost and nuclear norm minimization is elaborated along with PCA based constraint. The approach is evaluated for real noisy color images using standard datasets
in Section~\ref{sec:results}. Finally, the conclusion is drawn in Section~\ref{sec:conc}. 

\section{Related Works}
\label{sec:related}
The advent of SR techniques was started with multiple sub-pixel shifted LR images of the scene~\cite{overview}. Different sub-pixel shifted images are assumed to provide
different view points of the same scene. Hence, combining different view points 
can complement each other to produce an HR image~\cite{POCS}. Providing enough number
of images, the under-determined problem becomes determined to solve for the unknowns, and produce HR image~\cite{anr_SR03,Suresh_SR}. Further, these set of approaches are numerically limited to 
smaller factors~\cite{Glasner}. Additionally, the requirement of large number of images
became hindrance for such approaches. Hence, the focus of research has shifted towards
single image SR approaches, where the requirement of multiple LR images of the target
scene is replaced with the requirement of some HR example images~\cite{xample}. The assumption behind
such approaches is that the missing HF information of LR images can be imported from
HR example set. However, processing multiple HR images increases the memory requirement. Thus, patch based processing has been adopted for SR~\cite{yang,elad,ASDS,Aplus}. Further, there are efforts to super-resolve image as well as depth map jointly from LR stereo images~\cite{Bhavsar_TPAMI_SR}.

Absence of patches similar to the target patch in the database increases the complexity of the problem. This scenario can be addressed by including prior information about
the image. Natural image statistics such as smoothness prior has often been used in terms of Tikhonov, total variation, Markov random field, etc.~\cite{Tikhonov_sr,TVSR,MRF2,Bhavsar_CVIU_SR}. Among the other priors, non-local similarity and sparsity
inducing norms are the notable ones. Non-local similarity explores the patch similarity that are not constrained to a local region~\cite{Mairal,ASDS,Glasner}.
Sparsity inducing norm has been employed in SR approaches
on the basis that natural image is sparse in some domain~\cite{yangj,elad,ASDS,Mandal_SRSI}. While using sparsity inducing norm, the target
patch is generally represented by linear combination of few patches from the database
of patches, represented as columns of an over-complete matrix, known as dictionary~\cite{K-SVD,Mandal_SSD,Mandal_TIP}. Dictionary can have analytic form such as DCT or it can be learned from example patches~\cite{dict,yang,K-SVD,Mandal_SSD}. Sparsity prior
has often been combined with others such as non-local self similarity to improve SR performance~\cite{NCSR}. Most of these approaches will not work in the scenario when
example images are unavailable. Moreover, these approaches do not consider channel varying noise
in the models.

Surge of recent deep learning techniques has inspired researchers to employ deep convolutional neural network (CNN) for SR~\cite{SRCNN,accl_srcnn, Purohit18SR_ECCVW,vasu18_SR}. Since then, different CNN architectures have been employed for SR. Residual network~\cite{ResNet} has been
used to create deeper framework in conjunction with skip connection and recursive
convolution to improve the results~\cite{Kim_VDCNN,Kim_DRCN}. Nested skip connection 
has also been engaged with encoder-decoder architecture to improve convergence~\cite{Mao_NIPS16}. Most of these approaches appraise bicubic interpolated version of the LR image as input to the network~\cite{SRCNN,Kim_VDCNN,Kim_DRCN}. Processing
a higher dimensional image for very large number of levels requires higher computational resources. In order to avoid such condition, up-sampling module has
been appended at the end of the network~\cite{accl_srcnn,photo_GANSR,subpixel_CNN}.
However, these approaches can not deal with different scales~\cite{Kim_VDCNN}.
VDSR~\cite{Kim_VDCNN} has capability of training joint SR for different scales, and produce superior results than scale-specific network with the cost of higher computational burden. This requirement has been mitigated by using ResNet architecture
in the SRResNet model~\cite{photo_GANSR}. The ResNet architecture was originally proposed for different higher level vision tasks~\cite{ResNet}. Hence, direct application of it to SR may not be optimal. An optimized and simplified version of the
SRResNet has been proposed via EDSR to improve the results~\cite{EDSR}. To accommodate multiple
scales, MDSR proposed a multi-scale architecture that shares the set of parameters 
across different scales~\cite{EDSR}. The performance of all these deep learning based approaches
depends on availability of large number of example images. Moreover, the 
training-testing condition for these methods needs to be same. 

In order to alleviate such hard restrictions, small image-specific CNN has been developed based on patch recurrences in the input image~\cite{ZSSR}. The patch recurrence concept
has also been explored by some traditional approaches~\cite{Glasner,self_similar,epsr3,Icvgip14,SISR_noise,Trans_self_exem}. 
These methods super-resolve the given LR image based on the presumption that the HF information can be found out from similar patches across different scales. However, most of these techniques ignore the correlation among
the color channels. Further, the presence of noise in the LR image has often been neglected in the model.

The task of SR from a noisy LR image has been performed by a few techniques~\cite{SISR_noise,Mandal_SRSI,NNM_SR}. The approach~\cite{SISR_noise} poses the 
problem as a combination of denoising and SR. The noisy LR image is super resolved
directly to produce an HR version, which is conjugated with another HR image, derived
from denoised LR image in order to produce the final HR result. The motivation behind
the method is that the denoised LR image often lacks the HF information, which can be
imported from the super-resolved noisy HR image. The main drawback of it is that the
performance depends on the denoising algorithm. This issue has been taken care by implicit denoising while performing SR by the approach~\cite{Mandal_SRSI}. This
method estimates few parameters that are related to the noise statistics and are used in considering non-local mean or detail component for further processing.
However, the approach~\cite{Mandal_SRSI} considers only the luminance
component for SR. Thus, it can not take care of the channel varying noise. This holds
true for the approach~\cite{NNM_SR}, which applies nuclear norm minimization on 
luminance channel. In contrast,
we jointly super resolve the color channels by considering different weights for
different channels. Moreover, we consider weighted nuclear norm regularization to 
prioritize the significant singular values. Further, the 
solution is regularized by including a constraint that includes multi-scale 
image details through projection onto PCA basis.

The contributions of the proposed approach are summarized as follows:
\begin{itemize}
\item We propose to super-resolve color images in real noise by considering color 
channels jointly.
\item Different weights for different channels are estimated from their noise statistics, which are derived blindly from the input LR image. The estimated weights
are used in the data cost.
\item Nuclear norm minimization is employed with adaptive weights, which are assigned
based on significance of singular values. The weighted nuclear norm forms a
regularization term in our cost.
\item Multi-scale image details are augmented in the model as another
regularization term based on PCA and the combined objective function is
optimized using ADMM algorithm.
\end{itemize}

\section{Proposed Approach}
\label{sec:proposed}
We analyze noise statistics for color images in real noise to demonstrate it's channel
varying nature. This behavior is included in constructing the data cost function, which
is assisted with weighted nuclear norm and PCA basis based regularization term to model the problem. The entire cost function is optimized using 
alternating direction method of multipliers~\cite{ADMM}. 

\subsection{Inter-Channel Noise Statistics}
\label{subsec:noise_stat}
\begin{figure*}[!th]
\centering
\subfigure[]{\includegraphics[scale = 0.34]{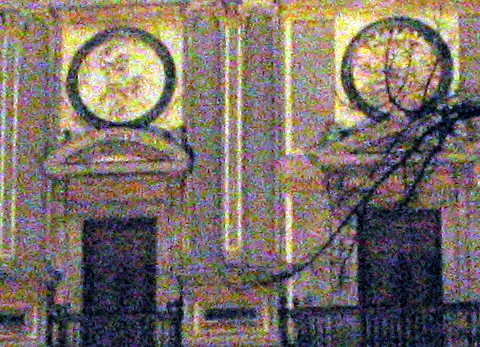}
       \label{fig:palace}
}
\subfigure[]{\includegraphics[scale = 0.32]{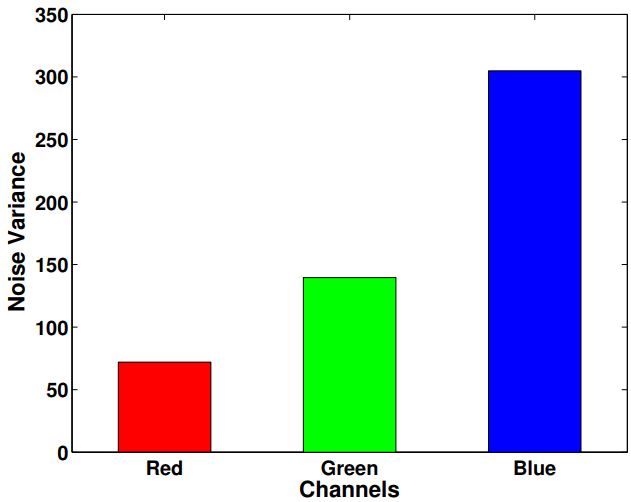}
       \label{fig:palaces}
}
\subfigure[]{\includegraphics[scale = 0.245]{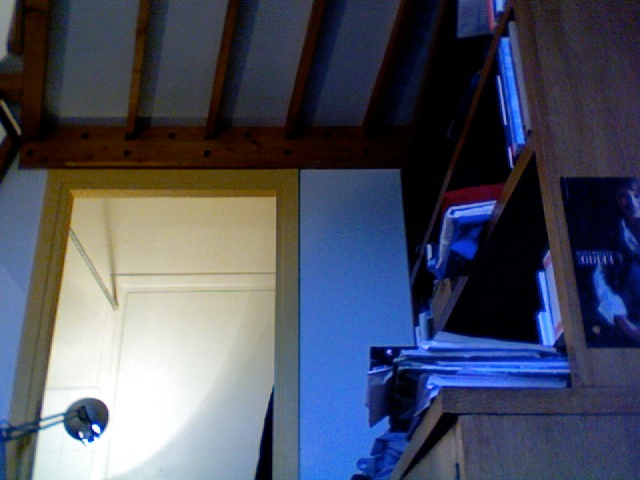}
       \label{fig:room}
}
\subfigure[]{\includegraphics[scale = 0.32]{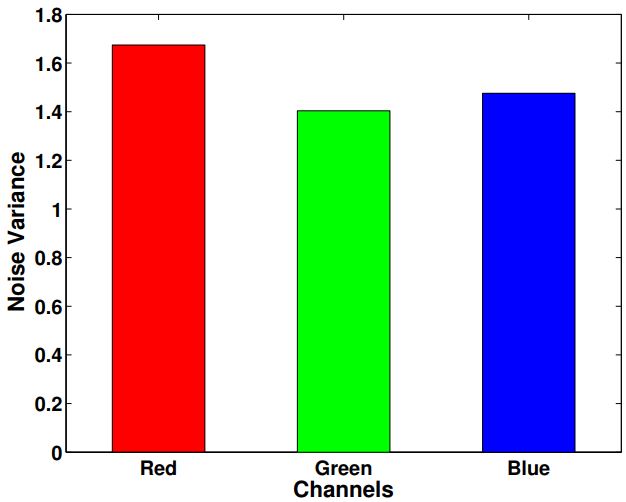}
       \label{fig:rooms}
}
\subfigure[]{\includegraphics[scale = 0.29]{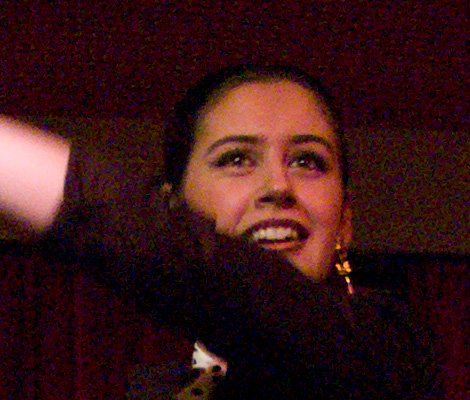}
       \label{fig:woman}
}
\subfigure[]{\includegraphics[scale = 0.32]{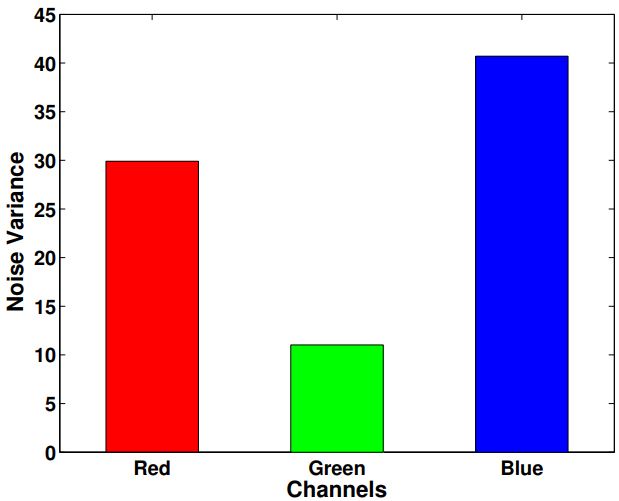}
       \label{fig:womans}
}
\vspace{-0.6cm}
\caption[]{Noise statistics across color channels for the images~\cite{real_data1}: (a) {\it Palace}; (c) {\it Room}; (e)  {\it Woman}. Corresponding variances of noise
for different channels are shown in (b,d,f), respectively.} 
\label{fig:noise_stat}
\end{figure*}

In order to check the behavior of noise across color channels, we estimate the noise
variances for real images using the technique~\cite{noise_var}. The real images are obtained from~\cite{real_data1}. The noise variances are shown in bar plots for different real examples in Fig.~\ref{fig:noise_stat}.
Note that the noise variances are not same across channels. Further, the variations 
across channels are also not uniform. Hence, depending on scenes, the {\it red, green \& blue} channels will be affected differently. Thus, processing only luminance 
component of the image or applying SR on every channel uniformly may not be suitable
for all images. This urges for an SR approach that takes care of the issue, which we
are going to discuss next.

\subsection{Model Formulation}
For a color image, the eq.~(\ref{eq:SR_prob}) can be re-written as
\begin{eqnarray}
\mathbf{y}_l = \mathbf{DHx}_l + \mathbf{n}_l, \;\;l \in \{r, g, b\}.
\end{eqnarray}
The objective is to recover $\mathbf{x}_l$ for each of the {\it red, green \& blue}
channels, represented by $r, g, b$. In order to proceed, the LR color image is up-sampled by off-the-shelf
interpolation technique to achieve an initial HR approximation, i.e., $\mathbf{\hat{x}}_l = \left(\mathbf{y}_l\right)\uparrow_d$. Patches are extracted in overlapping manner from the initial version as $\mathbf{x}_{i,l} = \mathbf{P}_{i}\mathbf{\hat{x}}_l$, where $\mathbf{P}_{i}$ extracts $i^{th}$ patch $\mathbf{x}_{i,l} \in \Re^{m}$ from $l^{th}$ color channel. Now, the patches from each of the color
channels are stacked together to form a vector $\mathbf{x}_i = 
[\mathbf{x}_{i,r};\mathbf{x}_{i,g};\mathbf{x}_{i,b}] \in \Re^{3m}$. For each of these
patches, we search for similarity in the image in terms of $l_2$ distance, and select
$s$ similar patches. These are kept in column-wise mode to form a matrix $\mathbf{X}_i \in \Re^{3m \times s}$.

In order to remove the bias of DC value, we extract detail component by subtracting 
a weighted mean from the matrix $\mathbf{X}_i$ as
\begin{eqnarray}
\mathbf{X}_{i,r} = \mathbf{X}_i - \mathbf{X}_{i,m},
\end{eqnarray}
where, $\mathbf{X}_{i,m}$ is the weighted mean of similar patches, and is estimated as
\begin{eqnarray}
\label{eq:NLM}
\mathbf{X}_{i,m} = \mathcal{R}\left[\sum_{j=1}^{s}  \left\{\frac{1}{z} \exp{\left(-\frac{||\mathbf{x}_i - \mathbf{x}_{j}||_2^2}{h}\right)}\right\} \mathbf{x}_{j}\right].
\end{eqnarray}
$\mathcal{R}$ repeats the vector to form a matrix of size equal to $\mathbf{X}_i$. $z$ is normalizing constant, and $h$ controls the decay of the exponential. 

Consider $\mathbfcal{X}_i$ is the clean HR counterpart of $\mathbf{X}_i$. Since,
$\mathbfcal{X}_i$ will have similar clean patches, the rank of it will be lesser.
It follows that the rank of $\mathbfcal{X}_{i,r}$ will also be lower.
By defining rank as number of non-zero singular values i.e., $\mathcal{K}\left(\mathbfcal{X}_{i,r}\right) = \sum_k|| \sigma_k\left(\mathbfcal{X}_{i,r}\right)||_0$,
we can minimize the rank along with a data cost that imposes the data 
continuity.
Solving an equation with $l_0$ norm minimization is NP-hard.
Thus rank is often relaxed as $\sum_k|| \sigma_k\left(\mathbfcal{X}_{i,r}\right)||_1$, which
is also known as nuclear norm $||\mathbfcal{X}_{i,r}||_*$~\cite{WNNM}. Hence, the cost function can be written as
\begin{eqnarray}
\label{eq:nuclear}
\hat{\mathbfcal{X}}_{i,r} = \displaystyle \arg \min_{\mathbfcal{X}_{i,r}} \left\{|| \mathbf{X}_{i,r} - \mathbfcal{X}_{i,r} ||^2_F + ||\mathbfcal{X}_{i,r}||_* \right\}.
\end{eqnarray}
In order to tackle the channel varying noise, we include a weight matrix $\bm{\Lambda}$ in the data term. Further, above equation minimizes all the singular values uniformly, irrespective of their significance. Considering the importance of different
singular values, we assign weights to minimize them differently~\cite{WNNM}.
The weighted nuclear norm can be written as 
\begin{eqnarray}
||\mathbfcal{X}_{i,r}||_{w,*} = \sum_k || w_k\sigma_k\left(\mathbfcal{X}_{i,r}\right)||_1 = \sum_k w_k\sigma_k\left(\mathbfcal{X}_{i,r}\right)
\end{eqnarray}
Thus, with the weights, the eq.~(\ref{eq:nuclear}) can be written as
\begin{eqnarray}
\label{eq:wt_nuclear}
\mathbfcal{\hat{X}}_{i,r} = \displaystyle \arg \min_{\mathbfcal{X}_{i,r}} \left\{|| \bm{\Lambda} \mathbf{X}_{i,r} - \bm{\Lambda} \mathbfcal{X}_{i,r} ||^2_F + ||\mathbfcal{X}_{i,r}||_{w,*} \right\}.
\end{eqnarray}

\subsubsection{The Weight $\bm{\Lambda}$:}
The weights $\bm{\Lambda}$ can be estimated using a MAP framework, where
we want to maximize the probability of $\mathbfcal{X}_{i,r}$, given $\mathbf{X}_{i,r}$ and $w$.
\begin{eqnarray}
\label{eq:map}
\mathbfcal{\hat{X}}_{i,r} &=& \displaystyle \arg \max_{\mathbfcal{X}_{i,r}} \left\{\ln P\left(\mathbfcal{X}_{i,r}|\mathbf{X}_{i,r},w\right)\right\}\nonumber \\ 
&=& \displaystyle \arg \min_{\mathbfcal{X}_{i,r}} \left\{-\ln P\left(\mathbf{X}_{i,r}|\mathbfcal{X}_{i,r}\right)- \ln P\left(\mathbfcal{X}_{i,r}|w\right)\right\}
\end{eqnarray}
Here, the term $P\left(\mathbf{X}_{i,r}|\mathbfcal{X}_{i,r}\right)$ is approximated 
by the noise statistics. According to our observation from Fig.~\ref{fig:noise_stat},
we can assume that noise is independent and identically distributed across channel.
Further, the distribution can be assumed to be Gaussian~\cite{CCINM}. Hence,
\begin{eqnarray}
P\left(\mathbf{X}_{i,r}|\mathbfcal{X}_{i,r}\right) = \prod_{l} \frac{1}{\left( 2\pi \sigma_l^2\right)^{\frac{3m}{2}}} \exp \left(-\frac{1}{2\sigma_l^2}
|| \mathbf{X}_{i,r,l} - \mathbfcal{X}_{i,r,l}||_F^2 \right).
\end{eqnarray}
The weighted nuclear norm is acted on $\mathbfcal{X}_{i,r}$. Hence, the term
$P\left(\mathbfcal{X}_{i,r}|w\right)$ will be proportional to  
$\exp\left(-\frac{1}{2} ||\mathbfcal{X}_{i,r}||_{w,*}\right)$. Putting these two terms in eq.~(\ref{eq:map}), we get
\begin{eqnarray}
\label{eq:combinem}
\hat{\mathbfcal{X}}_{i,r} = \displaystyle \arg \min_{\mathbfcal{X}_{i,r}} \sum_l || \frac{1}{\sigma_l} \mathbf{X}_{i,r,l} - \frac{1}{\sigma_l}\mathbfcal{X}_{i,r,l} ||^2_F + ||\mathbfcal{X}_{i,r}||_{w,*}.
\end{eqnarray}
Comparing, eq.~(\ref{eq:combinem}) with eq.~(\ref{eq:wt_nuclear}), we can write the
weight matrix $\bm{\Lambda}\in \Re^{3m\times 3m}$ as a diagonal matrix, whose non-zero entries are $1/\sigma_l$.
\subsubsection{The Weight $w$:}
The weight $w$ is assigned according to the significance of the singular values. For 
natural images, the larger singular values represents more important information than the
smaller ones. Hence, the larger singular values should be penalized lesser than the 
smaller singular values. Thus, one natural choice is to take inverse of the singular
values in some proportion. Here, we choose $w$ as~\cite{WNNM} 
\begin{eqnarray}
\label{eq:NN_wt}
w_k = \frac{C}{\sigma_k\left(\mathbfcal{X}_{i,r}\right) + \epsilon}, 
\end{eqnarray}
where $\sigma_k\left(\mathbfcal{X}_{i,r}\right)$ is the $k^{th}$ singular value of $\mathbfcal{X}_{i,r}$, $C$ and $\epsilon$ are constants.
\subsection{Employing Multi-Scale Image Details}
Here, we bring out the multi-scale image details in form of PCA basis. For 
a target patch $\mathbf{x}_i \in \Re^{3m}$, we find its similar patches 
across scales. The similar patches extracted from different up-scaled versions as well as down-scaled versions can provide the required patch
details to generate HR patch. This is because we perceive a coarser view 
of a scene from a long distance, and details of the scene reveal gradually
when we approach towards it. These similar patches are gathered together
in a column-wise manner to generate a matrix $\mathbf{T}_i$, which is
mean subtracted to unveil the details of different scales. The mean
subtracted matrix $\mathbf{T}_{i,r}$ is then used to find eigenvectors, which are further arranged in descending order corresponding to eigenvalues and placed in a matrix $\mathbf{B}_i$, which forms the required basis. 

The information, embedded in basis $\mathbf{B}_i$ is included into our formulation in form of a regularization term. Precisely, we project 
$\mathbf{X}_{i,r}$ onto $\mathbf{B}_i$, and revert it back via soft-thresholding and multiplication with basis. This should be closer to the
the matrix $\mathbfcal{X}_{i,r}$. Thus, the eq.~(\ref{eq:combinem}) is modified with the new term as
\begin{eqnarray}
\label{eq:pca}
\hat{\mathbfcal{X}}_{i,r} &=& \displaystyle \arg \min_{\mathbfcal{X}_{i,r}} \left\{ || \bm{\Lambda} \left(\mathbf{X}_{i,r} - \mathbfcal{X}_{i,r}\right) ||^2_F  + ||\mathbfcal{X}_{i,r}||_{w,*} + ||\mathbfcal{X}_{i,r} - \mathbf{B}_i\mathbfcal{S}\left(\mathbf{B}_i^T\mathbf{X}_{i,r}\right)_\alpha||^2_F \right\},
\end{eqnarray}
where, $\mathbfcal{S}\left(\cdot\right)_\alpha$ is a soft thresholding 
operator that shrinks the larger projection coefficients towards the center, and smaller coefficients to zero based on threshold $\alpha$.
\subsection{Optimization}
Unfortunately, the problem of eq.~(\ref{eq:pca}) does not have a closed
form solution because of the weight matrix $\bm{\Lambda}$ and the soft thresholding operator $\mathbfcal{S}$. In order to solve the equation, 
we introduce an augmented variable $\mathbfcal{F}$, which is used to 
represent the equation as a linear equality-constrained problem using
variable splitting method. To simplify expressions, in the optimization steps, we remove the subscripts of the variables. For example, $\mathbfcal{X}_{i,r}$ will be represented as $\mathbfcal{X}$. Hence, the cost can be written as
\begin{eqnarray}
\label{eq:aug}
\min_{\mathbfcal{X},\mathbfcal{F}} ||\bm{\Lambda} \left(\mathbf{X} - \mathbfcal{X}\right)||^2_F + 
||\mathbfcal{X} - \mathbf{B}\mathbfcal{S}\left(\mathbf{B}^T\mathbf{X}\right)_\alpha||^2_F + ||\mathbfcal{F}||_{w,*} \;\; s.t.\;\; \mathbfcal{X} = \mathbfcal{F}.
\end{eqnarray}
The eq.~(\ref{eq:aug}) is separable, hence it can be solved by 
alternating direction method of multipliers (ADMM)~\cite{ADMM}.
The augmented Lagrangian function becomes
\begin{eqnarray}
\label{eq:Lag}
\mathcal{L}\left(\mathbfcal{X},\mathbfcal{F},\bm{\Gamma}, \rho\right) = 
||\bm{\Lambda} \left(\mathbf{X} - \mathbfcal{X}\right)||^2_F + 
||\mathbfcal{X} - \mathbf{B}\mathbfcal{S}\left(\mathbf{B}^T\mathbf{X}\right)_\alpha||^2_F + ||\mathbfcal{F}||_{w,*} + \langle \bm{\Gamma}, \mathbfcal{X}-\mathbfcal{F}\rangle + \frac{\rho}{2} ||\mathbfcal{X}-\mathbfcal{F}||_F^2,
\end{eqnarray}
where $\bm{\Gamma}$ is the augmented Lagrangian multiplier, and $\rho$ is
the penalty parameter. We denote $\mathbfcal{X}_k$, $\mathbfcal{F}_k$ and
$\bm{\Gamma}_k$ are the optimization variables and Lagrangian multiplier
at $k^{th}$ iteration. Initialization of the variables are done 
by assigning zero matrices to $\mathbfcal{X}_0$, $\mathbfcal{F}_0$ and
$\bm{\Gamma}_0$. The penalty parameter is assigned a small positive value.
By taking derivative of the augmented Lagrangian function with respect 
to $\mathbfcal{X}$ \& $\mathbfcal{F}$, and equating it to zero, we can
update the variables in following manner:
\begin{enumerate}
\item Update $\mathbfcal{X}$:
\begin{eqnarray}
\mathbfcal{X}_{k+1} = \displaystyle \arg \min_{\mathbfcal{X}} ||\bm{\Lambda} \left(\mathbf{X} - \mathbfcal{X}\right)||^2_F + 
||\mathbfcal{X} - \mathbf{B}\mathbfcal{S}\left(\mathbf{B}^T\mathbf{X}\right)_\alpha||^2_F 
+ \frac{\rho_k}{2} ||\mathbfcal{X} - \mathbfcal{F}_k + \rho_k^{-1} \bm{\Gamma}_k||_F^2. 
\end{eqnarray}
This has a closed form solution:
\begin{eqnarray}
\mathbfcal{X}_{k+1} &=& \left(\bm{\Lambda}^T\bm{\Lambda} + \mathbf{I} + \frac{\rho_k}{2} \mathbf{I} \right)^{-1} \left(\bm{\Lambda}^T\bm{\Lambda} \mathbf{X} + \mathbf{B}\mathbfcal{S}\left( \mathbf{B}^T\mathbf{X}\right)_\alpha  + \frac{\rho_k}{2} \mathbfcal{F}_k - \bm{\Gamma}_k \right).
\end{eqnarray}
The soft-thresholding operator $\mathbfcal{S}$ is defined as
\begin{eqnarray}
\mathbfcal{S}\left(\mathbf{B}^T\mathbf{X}\right)_\alpha = sign \left(\mathbf{B}^T\mathbf{X}\right) \left(\vert \mathbf{B}^T\mathbf{X}\vert - \alpha \right)_{+}, 
\end{eqnarray}
where $\left(\vert \mathbf{B}^T\mathbf{X}\vert - \alpha \right)_{+}$ is 
established as
\begin{eqnarray}
\left(\vert \mathbf{B}^T\mathbf{X}\vert - \alpha \right)_{+} = 
 \begin{cases} 0 &\mbox{if }  \vert \mathbf{B}^T\mathbf{X}\vert<\alpha\\
\vert \mathbf{B}^T\mathbf{X}\vert - \alpha & \mbox{if } \vert \mathbf{B}^T\mathbf{X}\vert>\alpha \end{cases}
\end{eqnarray}

\item Update $\mathbfcal{F}$:
\begin{eqnarray}
\mathbfcal{F}_{k+1} = \displaystyle \arg \min_{\mathbfcal{F}} \frac{\rho_k}{2} || \mathbfcal{F} - \left( \mathbfcal{X}_{k+1} + \rho_k^{-1} \bm{\Gamma}_k \right) ||_F^2 + ||\mathbfcal{F}||_{w,*}
\end{eqnarray}
Let SVD of $\left( \mathbfcal{X}_{k+1} + \rho_k^{-1} \bm{\Gamma}_k \right) = \mathbf{U}_k \bm{\Sigma}_k \mathbf{V}_k^T$. According to eq.~(\ref{eq:NN_wt}), the weights are inversely proportional to the singular values. It follows that $0\leq w_1 \leq w_2 \cdots \leq w_n$. 
The work~\cite{WNNM} suggests that the above form of equation has a closed
form solution for non-decreasing weights and the solution is
\begin{eqnarray}
\mathbfcal{F}_{k+1} = \mathbf{U}_k \mathbfcal{S}\left(\bm{\Sigma}_k\right)_{\frac{w_k}{2}} \mathbf{V}_k^T.
\end{eqnarray}

\item Update $\bm{\Gamma}$:
\begin{eqnarray}
\bm{\Gamma}_{k+1} = \bm{\Gamma}_{k} + \rho_k\left(\mathbfcal{X}_{k+1} - \mathbfcal{F}_{k+1} \right)
\end{eqnarray}

\item Update $\rho$:
\begin{eqnarray}
\rho_{k+1} = \eta * \rho_k \;\;\mbox{where}\; \eta>1
\end{eqnarray}
\end{enumerate}
These steps are repeated until the algorithm converges or reaches the
maximum number of iterations.   
\subsection{Obtaining the Final Result}
The solution $\mathbfcal{X}_{i,r}$ is added with $\mathbf{X}_{i,m}$ of eq.~(\ref{eq:NLM}) to get the colored version $\mathbf{\hat{X}}_{i}$.
Columns of $\mathbf{\hat{X}}_{i}$ contains the super-resolved patches 
similar to $\mathbf{x}_i$. In this manner, all the patches are super-resolved ($\hat{\mathbf{x}}_{i,l}$), and are stitched together to form a full image $\hat{\tilde{\mathbf{x}}}_l$. Extracted patches from the full image should be consistent with the restored patches $\hat{\mathbf{x}}_{i,l}$. Further,
the recovered
HR image should be harmonious with the input image, if down-sampled in same
way. We need to get solution, which follows above two constraints, and
can be achieved by solving 
\begin{eqnarray}
 \label{eq:restore}
 \hat{\hat{\mathbf{x}}}_l = \displaystyle \arg \min_{\hat{\tilde{\mathbf{x}}}_l} \left\{ \sum_i || \mathbf{P}_i \hat{\tilde{\mathbf{x}}}_l - \hat{\mathbf{x}}_{i,l}||_2^2 
 + \beta \; || \mathbf{y}_l - \mathbf{DH}\hat{\tilde{\mathbf{x}}}_l ||_2^2 \right\}.
\end{eqnarray}
The first term recovers the entire image for each color channel from the 
recovered patches, and the second term is data continuity term. A closed form solution can be derived from eq.~(\ref{eq:restore})
\begin{eqnarray}
 \hat{\hat{\mathbf{x}}}_l = \left(\sum_i\mathbf{P}_i^T  \mathbf{P}_i + \beta \; \mathbf{H}^T\mathbf{D}^T\mathbf{DH} \right)^{-1} 
 \left( \sum_i \mathbf{P}_i^T \hat{\mathbf{x}}_{i,l} +  \beta \;\mathbf{H}^T\mathbf{D}^T \mathbf{y}_l \right).
\end{eqnarray}
The recovered $\hat{\hat{\mathbf{x}}}_l$ for each color channel can be 
spliced together to produce the final color image.

\section{Experimental Results}
\label{sec:results}
We evaluate the proposed approach by considering real images as provided
by~\cite{CCINM,real_data1}. The dataset~\cite{CCINM} is created by capturing images of 11 static scenes under controlled indoor environment using different cameras with different settings.
However, a camera and its settings are kept fixed for shooting a particular scene. In this manner, 500 images per scene are captured. The mean of these
images can serve as ground truth for each of the scenes and can be used
for computing quantitative measurements such as PSNR, and SSIM. Originally, dimensions of the captured images are quite large (of the order $7000 \times 5000$). However, 15 cropped versions of the images with dimension $512 \times 512$ are provided by the authors of~\cite{CCINM}. These 15
smaller images are used in our approach for experimentation. In contrast,
the dataset~\cite{real_data1} is constructed in uncontrolled environment.
A set of 20 images of different dimensions are considered from the dataset~\cite{real_data1} for experimental purpose. Absence of ground truths in the dataset restricts us to do only visual comparison.

The noisy images are down-sampled by factor 3 using MATLAB command $imresize$ with $bicubic$ interpolation to generate LR images. The noise 
statistics for each color channels are estimated from the LR images using~\cite{noise_var}. The LR image is then up-scaled using bicubic 
interpolation technique to generate an initial approximation of HR image. 
Patches of size $6 \times 6$ are extracted from the initial HR image. For
each patch, we search for its similarity in $25 \times 25$ neighborhood and
consider 20 most similar patches. To accommodate image details in the 
model, we sample the initial HR image into 6 different levels by factors $(0.8)^i$, where $i = 1,2,3,4,5,6$. The initial value of $\rho$ is set to 
be 1. The $\alpha$ for soft-thresholding operator is chosen as 0.8. The maximum iterations are limited to 360.  
The computational complexity of this approach is $\mathcal{O}(m^3 L)$ , where $m$ is the size of a patch, and $L$ is the number of patches.
We have compared our results with various approaches including conventional
approaches~\cite{yang,ASDS,elad,Aplus,SPSR,Mandal_SRSI,NNM_SR} as well as deep learning based approaches\cite{SRCNN,EDSR,ZSSR}. 

\begin{figure*}
	\centering
		\begin{tabular}{c@{ }  c@{ } c@{ }  c@{ } c@{ }}	
		\hspace{-0cm}\includegraphics[trim = 0 0 0 0, clip, width=80pt]{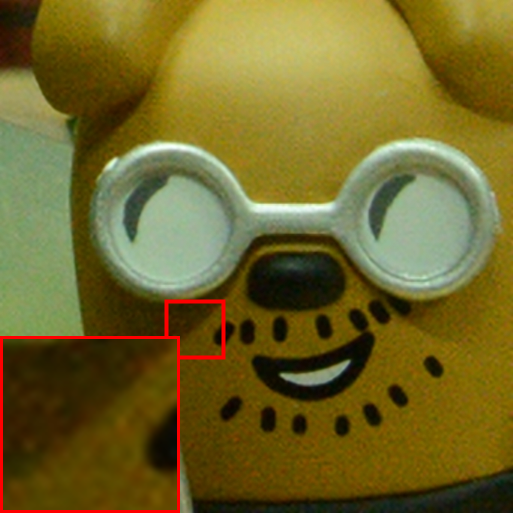}
		\hspace{-0cm}\includegraphics[trim = 0 0 0 0, clip, width=80pt]{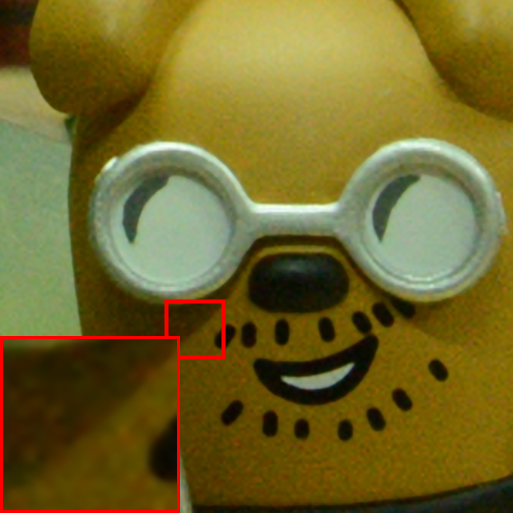}
		\hspace{-0cm}\includegraphics[trim = 0 0 0 0, clip, width=80pt]{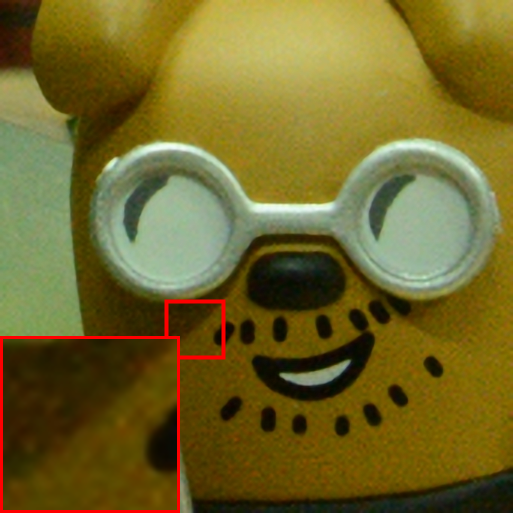}
		\hspace{-0cm}\includegraphics[trim = 0 0 0 0, clip, width=80pt]{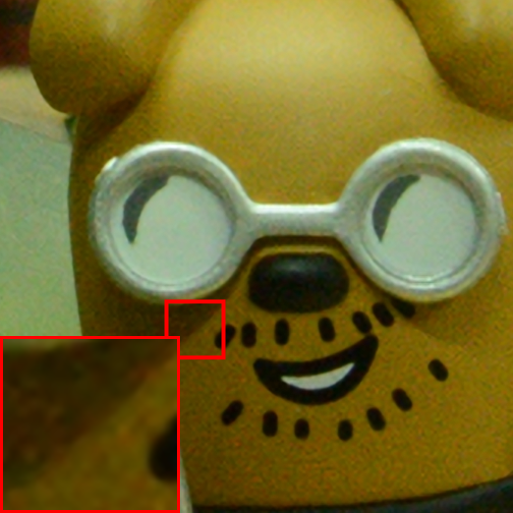}
		\hspace{-0cm}\includegraphics[trim = 0 0 0 0, clip, width=80pt]{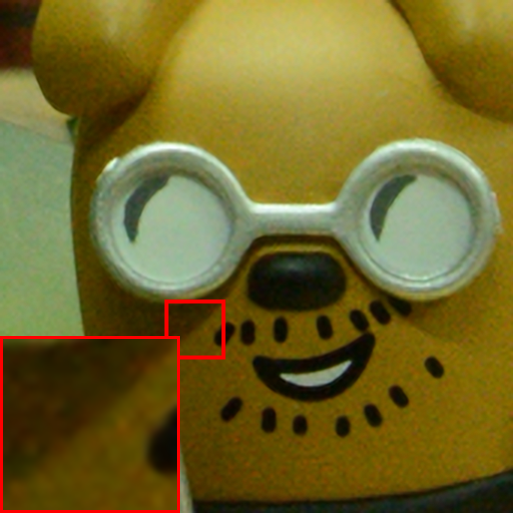}\\
		(a) \hspace{2.4cm} (b) \hspace{2.4cm} (c) \hspace{2.4cm} (d) \hspace{2.4cm} (e) \\
		\hspace{-0cm}\includegraphics[trim = 0 0 0 0, clip, width=80pt]{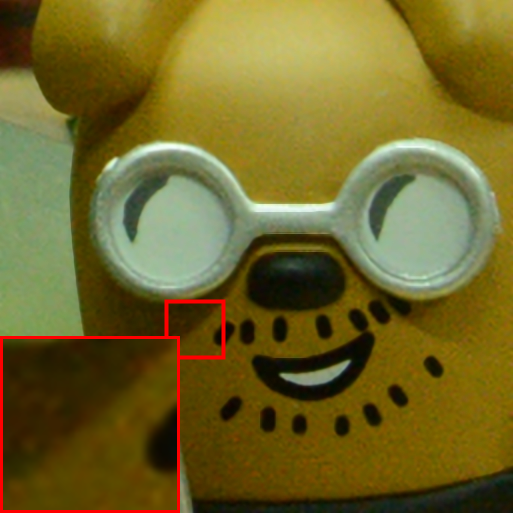}
		\hspace{-0cm}\includegraphics[trim = 0 0 0 0, clip, width=80pt]{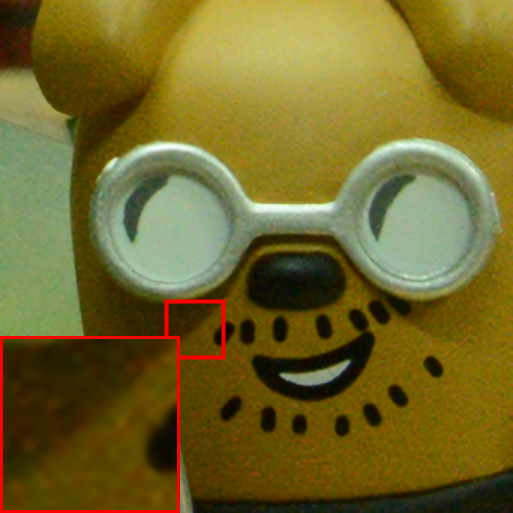}
		\hspace{-0cm}\includegraphics[trim = 0 0 0 0, clip, width=80pt]{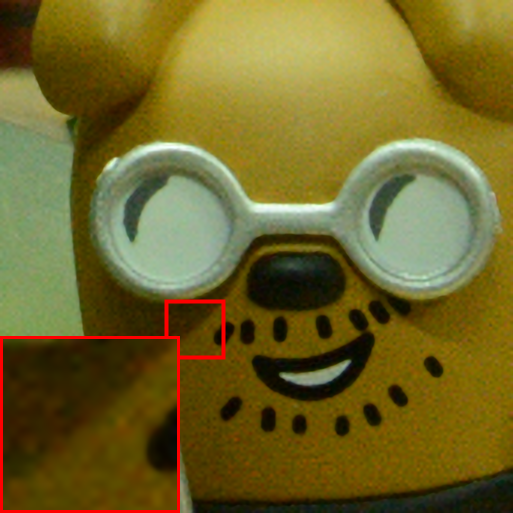}
		\hspace{-0cm}\includegraphics[trim = 0 0 0 0, clip, width=80pt]{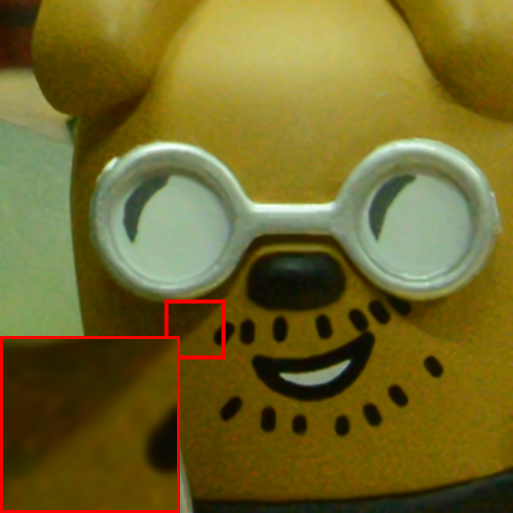}
		\hspace{-0cm}\includegraphics[trim = 0 0 0 0, clip, width=80pt]{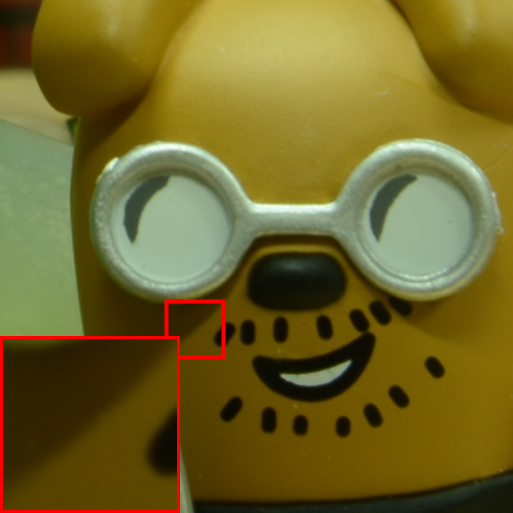}\\					
(f) \hspace{2.4cm} (g) \hspace{2.4cm} (h) \hspace{2.4cm} (i) \hspace{2.4cm} (j) \\
		\end{tabular}
		\vspace{-0.5em}	
		\caption{\label{fig:result} Visual results on $1^{st}$ scene of Nikon D800 (ISO = 3200) from Table~\ref{tab:gt}. (a)-(i)
		depict the results of ASDS\cite{ASDS}, SU\cite{elad}, SRCNN\cite{SRCNN}, A+\cite{Aplus}, SPSR\cite{SPSR}, NASR\cite{Mandal_SRSI}, EDSR\cite{EDSR}, ZSSR\cite{ZSSR}, \& our result, respectively; (j) represents the Ground truth.}
\end{figure*}

\subsection{Experiments Using Dataset~\cite{CCINM}}
The quantitative measures such as PSNR and SSIM\footnote{PSNR values are positioned above the SSIM values in each cell of the table.} for the results on 15 images of the dataset~\cite{CCINM} are depicted in Table~\ref{tab:gt} along with the results of existing approaches.
\begin{table*}
\centering
\caption{Results of SR on dataset~\cite{CCINM}($\uparrow 3$), captured by Nikon D800}
\vspace{-4mm}
\label{tab:gt}
\resizebox{\columnwidth}{!}{
\begin{tabular}{|l|c|c|c|c|c|c|c|c|c|c|}
\hline
ISO & RP\cite{yang} & ASDS\cite{ASDS} & SU\cite{elad} & SRCNN \cite{SRCNN} & A+\cite{Aplus} &SPSR\cite{SPSR} & NASR\cite{Mandal_SRSI}& EDSR \cite{EDSR} & ZSSR\cite{ZSSR} & ours\\
\hline
\hline
& 40.35 & 40.70 & 41.15 & 40.81 & 41.12 & 41.38 & 40.91 & 41.40 &40.52 & {\bf 41.65} \\
   & 0.9684 & 0.9644 & 0.9694 & 0.9655 & 0.9676 & 0.9720 & 0.9726 & 0.9691 & 0.9654 & {\bf 0.9741} \\
\cline{2-11}
  & 37.05 & 37.22 & 37.78 &38.13 & 37.40 & 35.57 & 37.95 & 37.98 &36.82 & {\bf 38.78 }\\
 1600   & 0.9761 & 0.9640 & 0.9694 & 0.9658 & 0.9652 & 0.9678 & 0.9754 & 0.9677 & 0.9614 & {\bf 0.9785}\\  
\cline{2-11} 
 & 40.68 & 40.20 & 40.64 & 40.25 & 40.59 & 40.88 & 41.06 & 40.74 &39.36 & {\bf 41.49 } \\
   & 0.9619 & 0.9506 & 0.9574 & 0.9520 & 0.9551 & 0.9616 & 0.9641 & 0.9560 & 0.9468 & {\bf 0.9657}\\  
\hline
& 40.95 & 39.43 & 39.96 &39.46 & 39.73 & 40.53 & 40.60 & 39.78 &39.66 & {\bf 41.75} \\
   & 0.9760 & 0.9468 & 0.9579 & 0.9487 & 0.9518 & 0.9642 & 0.9660 & 0.9511 & 0.9472 &{\bf 0.9752}\\
\cline{2-11}
 & 39.83 & 40.17 & 40.61 &40.29 & 40.42 & 40.89 & 40.42 & 40.66 &39.67 & {\bf 41.39} \\
3200   & 0.9676 & 0.9627 & 0.9683 & 0.9638 & 0.9653 & 0.9710 & 0.9649 & 0.9664 & 0.9627 & {\bf 0.9747}\\  
\cline{2-11} 
 & 40.62 & 39.27 & 40.07 & 39.35 & 39.53 & 40.86 & 40.92 & 39.53 & 39.25 & {\bf 42.53 }\\
   & 0.9320 & 0.9255 & 0.9416 & 0.9278 & 0.9309 & 0.9529 & 0.9540 & 0.9308 & 0.9276 & {\bf 0.9700}\\  
\hline
& 34.14 & 33.78 & 33.97 & 34.03 & 33.87 & 34.10 & {\bf 35.79} & 34.93 &33.74 & 35.49 \\
   & 0.9184 & 0.8682 & 0.8881 & 0.8733 & 0.8764 & 0.8970 & 0.9153 & 0.8832 & 0.8691 & {\bf 0.9226} \\
\cline{2-11}
  & 34.79 & 35.05 & 35.21 &35.08 & 35.24 & 35.39 & 35.57 & 35.49 &35.07 & {\bf 36.00} \\
6400   & 0.9416 & 0.9238 & 0.9334 & 0.9264 & 0.9285 & 0.9383 & 0.9455 & 0.9299 & 0.9230 & {\bf 0.9500}\\  
\cline{2-11} 
 & 35.31 & 34.91 & 35.40 &35.02 & 35.21 & 35.51 & 35.28 & 35.39 &34.12 & {\bf 36.21} \\
   & 0.9219 & 0.9023 & 0.9167 & 0.9059 & 0.9087 & 0.9192 & 0.9093 & 0.9110 & 0.8926 & {\bf 0.9330}\\  
\hline
\end{tabular}
}
\end{table*}
Among these approaches, RP\cite{yang}, ASDS\cite{ASDS}, SU\cite{elad},
SRCNN\cite{SRCNN}, Aplus\cite{Aplus}, SPSR\cite{SPSR}, EDSR\cite{EDSR} depends on example image patches. However, NASR\cite{Mandal_SRSI}, NNSR\cite{NNM_SR}, and ZSSR\cite{ZSSR} do not require example image patches. It can be observed from the table that for most of the cases, we are able to produce the best results, as denoted by bold fonts. Further, the superiority of our approach can be verified visually through an example, as shown in Fig.\ref{fig:result}.
From first glance, the results appear to be more or less similar.
However, difference can be found in the zoomed in part of the image.
The example image based approaches~\cite{ASDS,elad,SRCNN,Aplus,SPSR,EDSR} are not able to reduce the noisy
artifacts much as they have not seen images with real noise and their ground truths in the training set. Further,
these approaches do not model the noise with its channel varying 
characteristic. Though, NASR~\cite{Mandal_SRSI} can take care of the 
noise but the channel varying nature of it restricts the method from performing
well, as can be observed in (f). On the other side, ZSSR~\cite{ZSSR}
is a deep learning based approach, which over-fits on the given LR image
to produce an SR result, which is still infected by noise.
Whereas, we are able to reduce the noisy artifacts due to the elegant combination of weighted data cost, weighted nuclear norm
and multi-scale image details. 
\begin{figure*}[t]
	\centering
		\begin{tabular}{c@{ }  c@{ } c@{ }  c@{ } c@{ } c@{ }}	
		\hspace{-0cm}\includegraphics[trim = 0 0 0 0, clip, width=80pt]{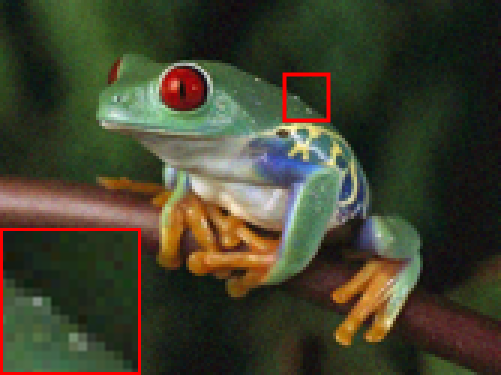}&
		\hspace{-0cm}\includegraphics[trim = 0 0 0 0, clip, width=80pt]{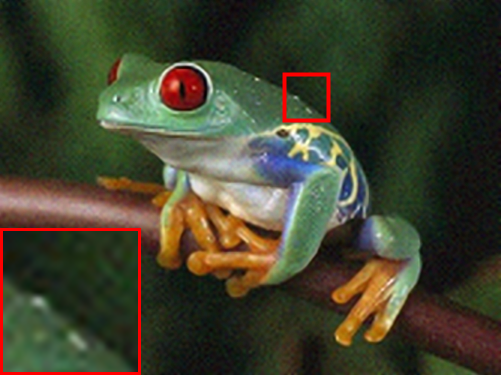}&
		\hspace{-0cm}\includegraphics[trim = 0 0 0 0, clip, width=80pt]{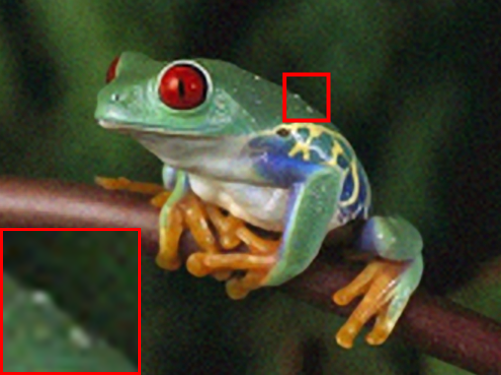}&
		\hspace{-0cm}\includegraphics[trim = 0 0 0 0, clip, width=80pt]{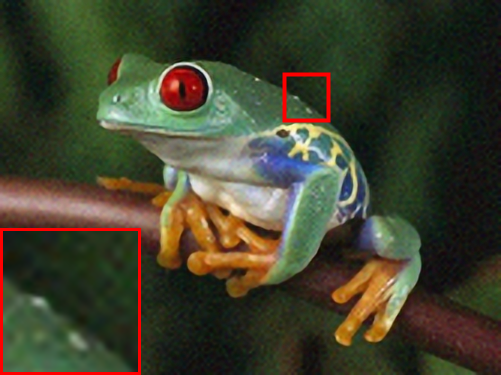}&
		\hspace{-0cm}\includegraphics[trim = 0 0 0 0, clip, width=80pt]{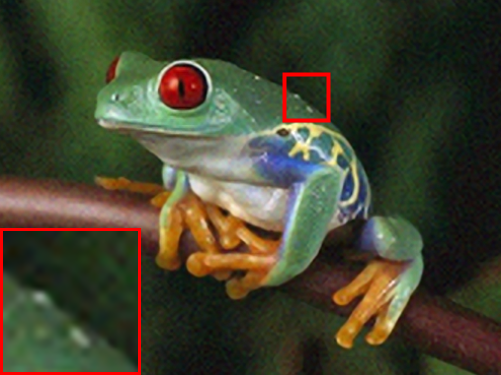} \\
		(a) & (b) & (c) & (d) & (e) \\
		\hspace{-0cm}\includegraphics[trim = 0 0 0 0, clip, width=80pt]{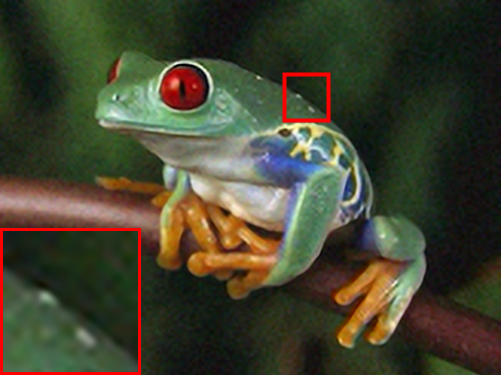}&
		\hspace{-0cm}\includegraphics[trim = 0 0 0 0, clip, width=80pt]{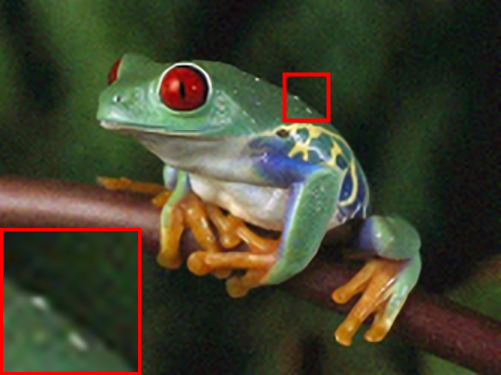}&
		\hspace{-0cm}\includegraphics[trim = 0 0 0 0, clip, width=80pt]{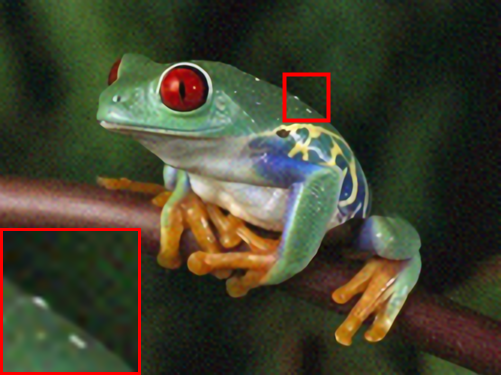}&
		\hspace{-0cm}\includegraphics[trim = 0 0 0 0, clip, width=80pt]{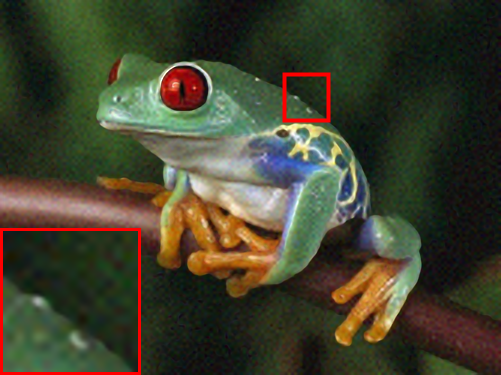}&
		\hspace{-0cm}\includegraphics[trim = 0 0 0 0, clip, width=80pt]{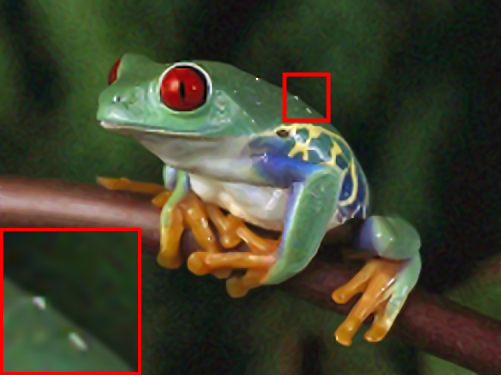}&
		\\			
		(f) & (g) & (h) & (i) & (j) \\
		\end{tabular}
		\vspace{-0.5em}	
		\caption{\label{fig:result1} Visual results on $Frog$ from the dataset~\cite{real_data1}. (a) shows the LR image, (b)-(j)
		depict the results of ASDS\cite{ASDS}, SU\cite{elad}, SRCNN\cite{SRCNN}, A+\cite{Aplus}, SPSR\cite{SPSR}, NASR\cite{Mandal_SRSI}, EDSR\cite{EDSR}, ZSSR\cite{ZSSR}, and our result, respectively.}
\end{figure*}
 
\subsection{Experiments Using Dataset~\cite{real_data1}}
This dataset does not contain any ground truth of the noisy observations. 
Hence, the computation of the metrics PSNR \& SSIM are not possible for the
dataset.
Here, we show the results of different approaches along with the results 
of our method in Fig.~\ref{fig:result1}.
One can observe that existing approaches including the deep learning ones are not able to reduce the noisy artifacts. Further, the edges of some of their results are smeared. However, our approach is able to suppress the effect of noise without smearing the edges. 

\section{Conclusion}
\label{sec:conc}
We proposed to super-resolve an image in real noise by alleviating the 
channel varying noise of it. Weights, adaptive to the noise statistics were assigned to the data cost, which was augmented by two regularization terms. One of the terms maintains the low-rank property of similar patches
by weighted nuclear norm minimization, where weight carries the significance of singular values. Multi-scale image details were embedded into
the model through the second regularization term, which was constructed
via projection onto PCA basis. The combined objective function was minimized using
ADMM-based optimization algorithm, which leads to suppression of noise while bringing out image details. 
The results demonstrated the super-resolving capability of our approach 
in real noise.

\vspace{-0.5cm}
\bibliographystyle{ACM-Reference-Format}
\bibliography{edge_sr}

\end{document}